\documentclass[runningheads,orivec]{llncs}
\usepackage{graphicx,amsmath,bm}
\usepackage[T1]{fontenc}
\usepackage{caption}
\usepackage[subrefformat=parens]{subcaption}
\usepackage{amsmath}
\usepackage{algorithm}
\usepackage{algorithmic}
\usepackage{comment}
\usepackage{float}
\usepackage{bm}
\begin{document}
\pagestyle{empty}
\title{Basins of Attraction to Multiple Fixed Points in Discrete-time Hysteresis Neural Networks}
%
\titlerunning{Basin of attraction}
%
\author{Yuta Arai \and 
Seigo Nakamura \and 
Ryoga Nakamura \and 
Muzuki Ohira \and  \\ 
Toshimichi Saito}
%
\authorrunning{Yuta Arai et. al. }
%
\institute{$^1$Hosei University, Koganei, Tokyo, Japan,  tsaito@hosei.ac.jp
\\
}
\maketitle              
\begin{abstract}
This paper studies multiple fixed points in a discrete-time hysteresis neural network. 
The network consists of binary hysteresis neurons characterized by the threshold parameter. 
Depending on the parameter, the network can have a variety of multiple binary fixed points. 
Stability of each fixed point is characterized by basin of attraction (BOA): the set of initial points falling into the fixed point. 
In order to evaluate the distribution of BOA sizes, we use the normalized entropy. 
In order to escape from the curse of dimensionality, we introduce a simple problem: classification of binary data set. 
In the classification, BOAs  correspond to classes. 
In the problem, we clarify that the threshold parameter can control the BOAs, especially, can maximize the entropy: the distribution approaches to uniform. 
As a concrete example, we consider an item response data set in education. 
Using two fundamental metrics in the item response theory, the classification results are evaluated. 
\end{abstract}
\section{Introduction}
Discrete-time recurrent neural networks are important/interesting dynamical systems and have been studied extensively \cite{hopf} \cite{amari} \cite{liu}. 
The networks are characterized by nonlinear activation function and connection parameters. 
The dynamics is described by an autonomous difference equation. 
Depending on the parameters, the networks can generate various nonlinear phenomenon: fixed points, periodic orbits, and chaos. 
The nonlinear phenomena can co-exist and the network exhibits one of them depending on initial condition. 
Analysis of the phenomena is an important problem in fundamental study of nonlinear dynamics. 
Among various nonlinear phenomena, we consider multiple fixed points. 
They are applicable to
associative memories \cite{amari} \cite{michel} \cite{araki}, 
error correcting codes \cite{error}, 
control systems \cite{control}, and so on. 
Stability analysis of the multiple fixed points is important from both fundamental and theoretical viewpoints. 
Various valuable results are published \cite{liu}, however, analysis is not sufficient in basin of attraction 
(BOA, the set of initial points leading to transient behavior that approaches the fixed point). 
However, analysis of the BOA is not easy. 
Even in small dynamical systems, the analysis is difficult \cite{ott}. 
In order to realize the analysis, simple models and problems are necessary. 

This paper studies the BOAs in a discrete-time hysteresis neural network (HYN \cite{jk1} \cite{jk2} \cite{jk3} \cite{nolta}, a simple model) in classification problem of binary data sets (a simple problem).  
The HYN consists of binary hysteresis neurons and has two kinds of parameters: hysteresis threshold and connection among neurons. 
The dynamics is described by a simple autonomous difference equation of binary state variables.  
Depending on the parameters, the HYN can generate a variety of binary fixed points. 
Although the HYN has many parameters, we select the hysteresis threshold as a control parameter: it is convenient to analyze the dynamics. 
Since each fixed point has its BOA, multiple fixed points correspond to the set of BOAs. 
The set of BOAs is characterized by distribution of the BOA sizes. 
In order to evaluate the distribution, we use normalized entropy. 

Using artificial data sets, we clarify that the hysteresis threshold can control the BOAs. 
Especially, it can maximize the entropy: the distribution approaches to uniform. 
If all of the multiple fixed points have the same importance, the uniform distribution is suitable.  

As a first step to applications, we consider capability of the HYN to classification of binary data sets in education \cite{fujita} \cite{kiyohara}. 
In the classification, multiple fixed points correspond to class centers and their BOAs correspond to classes. 
Using two fundamental metrics from item response theory (IRT \cite{irt}), the classification performance is evaluated.

\section{Discrete-time Hysteresis Neural Networks}

\begin{figure}[tb]
\centering
\includegraphics[width=0.5\columnwidth]{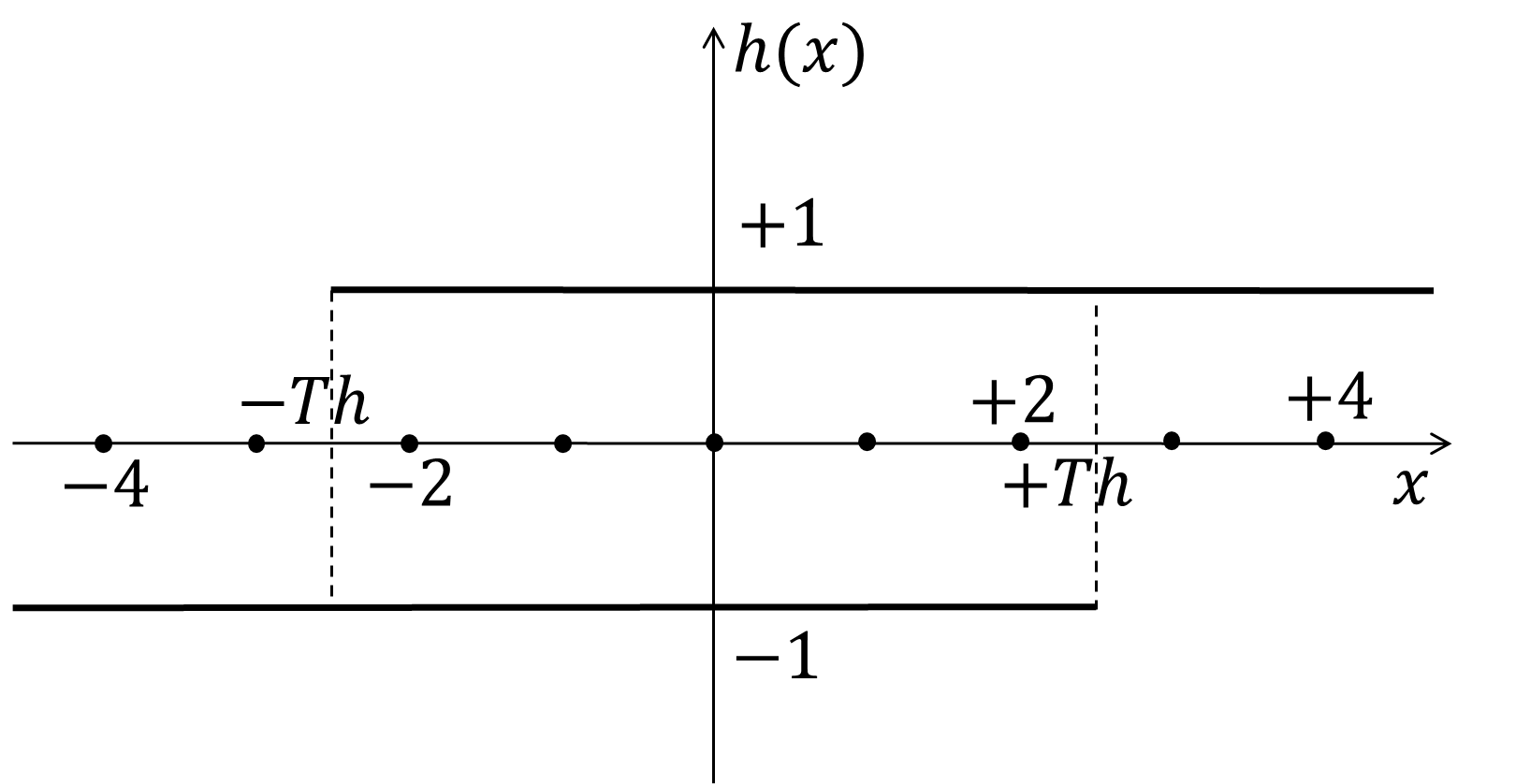}
\caption{Hysteresis Activation function.}
\label{fg1}
\end{figure}

The discrete-time hysteresis neural network (HYN) is described by autonomous difference equation of binary state variables:
\begin{equation}
    \label{hnn}
\begin{array}{l}
x_i(t+1) = h(y_i(t)), \ i \in \{1, \cdots, N \}\\ 
\displaystyle y_i(t)=\sum_{j=1}^{N} w_{ij} x_j(t)\\
h(y_i(t)) =
\begin{cases}
+1 & \text{for } y_i(t) \ge +Th \\
x_i(t) & \text{for } |y_i(t)| < |Th| \\
-1 & \text{for } y_i(t) \le -Th
\end{cases} \ \ \mbox{ for } Th>0
\end{array}
\end{equation}%
where $x_i(t) \in \{-1, +1\}$ is the $i$-th binary state variable at discrete time $t$. 
Weighted sum of $x_i(t)$ is applied to the hysteresis activation function $h$ as shown in Fig. \ref{fg1}. 
The connection parameters $w_{ij}$ are integers and the hysteresis threshold parameter is 
\[
Th \in \{ 0.5, 1.5, 2.5, 3.5, \cdots  \}
\]
Since input $y_i(t)$ is an integer, 
$Th$ has the margin $0 \le d < 0.5$. 
If $Th=0$, the hysteresis activation function becomes the signum activation function
\begin{equation}
h(y_i(t)) = \mbox{sgn}(y_i(t)) =
\begin{cases}
+1 & \text{for } y_i(t) \ge 0 \\
-1 & \text{for } y_i(t) < 0
\end{cases} \ \ \mbox{ for }  Th=0\\
\end{equation}
The threshold parameter $Th$ plays crucial roles in behavior of the HYN.  
For convenience, we introduce the vector form
\begin{equation}
    \bm{x}(t+1) = F(\bm{W} \bm{x}(t)), \ \
\bm{W} \equiv
\begin{pmatrix}
w_{11} & \cdots & w_{1N}\\
\vdots & \ddots & \vdots\\
w_{N1} & \cdots & w_{NN}\\
\end{pmatrix}, \  \
\bm{x} \equiv 
\begin{pmatrix}
x_1\\
\vdots\\
x_N
\end{pmatrix} 
\end{equation}
where connection parameters are summarized into the connection matrix
$\bm{W}$. 

\section{Fundamental Concepts}

As preparations to analyze the HYN, we consider/introduce several concepts. 

\subsection{Basin of attraction to fixed points}

\begin{figure}[tb]
\centering
\includegraphics[width=0.5\columnwidth]{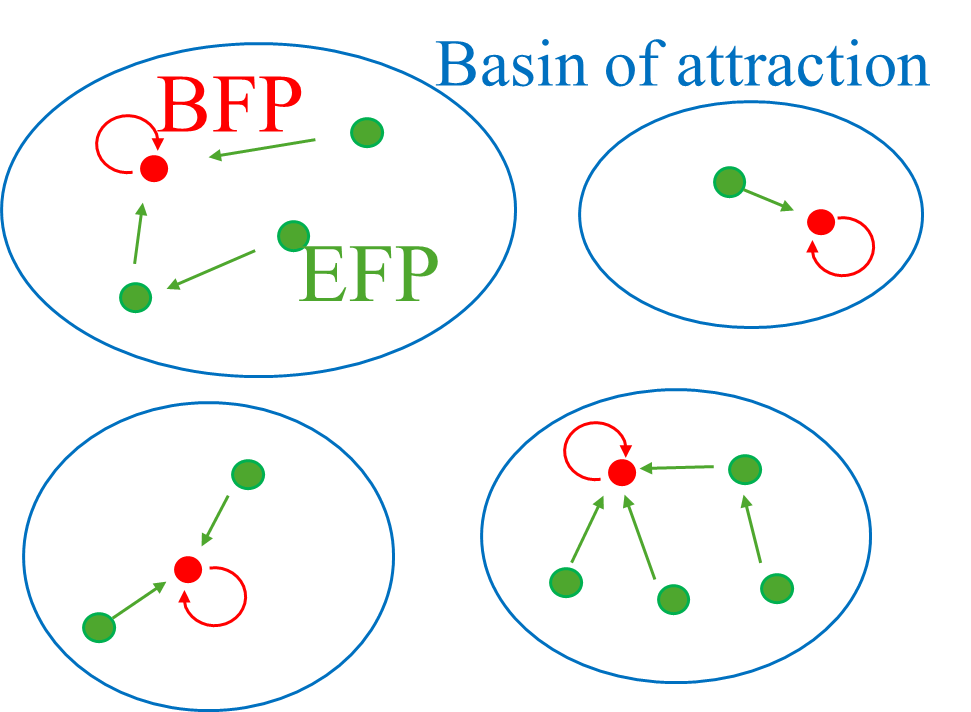}
\caption{Binary fixed points (BFPs), eventually fixed points (EFPs), and basin of attraction (BOA).}
\label{efp}

\vspace*{3mm}

\centering
\includegraphics[width=0.8\columnwidth]{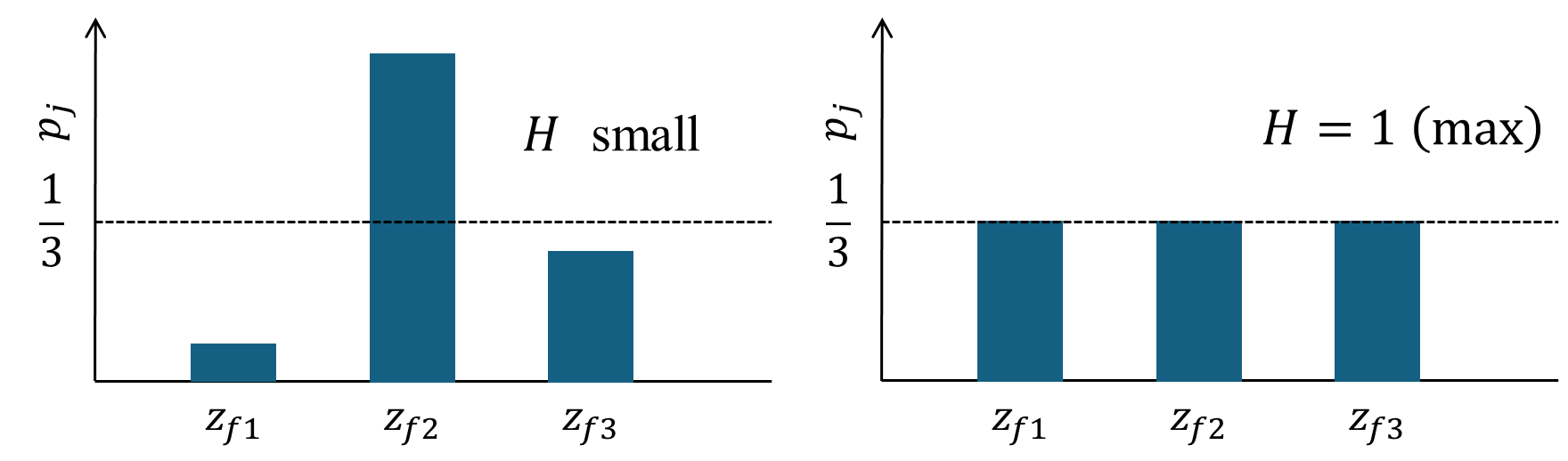}
\caption{Normalized distribution of BOA sizes. 
($z_{f1}$, $z_{f2}$, $z_{f3}$): BFPs. 
$H$: normalized entropy.}
\label{entropy}
\end{figure}

First, we give several definitions. 
A binary vector $\bm{z}_f$ is said to be a binary fixed point (BFP) if 
\[
F(\bm{W}\bm{z}_f)=\bm{z}_f. 
\]
A binary vector $\bm{z}_e$ is said to be an eventually fixed point (EFP) if 
a sequence started from $\bm{z}_e$ falls into some fixed point $\bm{z}_f$:
\[
F^l(\bm{W} \bm{z}_e) = \bm{z}_f, \  \bm{z}_e \ne \bm{z}_f \  
\mbox{ for some } l
\]
where $F^l$ is the $l$-fold composition of $F$ and $l$ is some positive integer. 
Fig. \ref{efp} illustrates BFPs and EFPs. 

Let an HYN has $m$ fixed points
$\bm{z}_{f1}, \cdots, \bm{z}_{fm}$. Let 
the $j$-th fixed point $\bm{z}_{fj}$ have $k_j$ EFPs. 
We define the set of the EFPs of the $j$-th BFP: 
\[
A_j = \{\bm{z}_{e1}, \cdots. \bm{z}_{ek_j}, \bm{z}_{fj}  \}, \   
F^l(\bm{W} \bm{z}_{ei}) = \bm{z}_{fj}
\mbox{ for } i \in \{1, \cdots, k_j \}, \ j \in \{1, \cdots, m\}. 
\]
The set $A_j$ is referred to as the basin of attraction (BOA) to the $j$-th fixed point $\bm{z}_{fj}$. 
For convenience in analysis, the fixed point $z_{fj}$ is added to the $A_j$. 
$k_j$ is the number of EFPs, 
that characterizes the size of $A_j$. 
Then we define
\begin{equation}
\begin{array}{l}
\mbox{The set of BOAs: } A \equiv \{A_1, \cdots, A_m \}\\
\mbox{Distribution of BOA sizes: }
K \equiv \{k_1+1, \cdots, k_m+1 \}
\end{array}
\label{dist}
\end{equation}
where $A_j$ consists of $k_j$ EFPs.  
The distribution $K$ 
is fundamental in the analysis of BOAs. 
As a feature quantity of the distribution, we use normalized entropy
\begin{equation}
H = \frac{H_m}{H_{max}},  \ 
H_m = - \sum_{j=1}^{m} p_j \log_2 p_j, \
p_j = \frac{k_j}{k_1 + \cdots + k_m},  \ 
H_{max} = \log_2 m
\end{equation}
where $\{p_1, \cdots, p_m\}$ is the normalized distribution corresponding to probability function. 
As the distribution approaches uniform, $p_1 = \cdots = p_m =1/m$, the entropy approaches 1. 
Fig. \ref{entropy} illustrates normalized distribution and entropy. 
The entropy is used to characterize/evaluate the BOAs. 
If the entropy is the maximum ($H=1$), the BOAs have the same size. 
It is convenient if uniform classification is required. 

\subsection{Associative memories}
As stated earlier, the recurrent neural networks have been applied to a variety of systems. 
Among them, the associative memory is one of important/interesting applications. 
We briefly introduce associative memories in discrete-time recurrent neural networks with signum activation function.  
First, desired memories, several binary vectors, are given. 
Second, applying some learning method, we try to store the desired memories as fixed points. 
In the storage, the problem is efficient setting of the connection parameters $w_{ij}$ to store the desired memories. 
Third, if desired memories are stored,  the stability of desired memories are investigated.  
In historical/important works, 
fundamental results are given. 
In Ref. \cite{amari}, applying the  statistical neuro-dynamics theory, memory capacity is investigated. 
Roughly speaking, if the number of desired memories is at most $0.15N$, 
the desired memories can be stored for sufficiently large $N$. 
In Ref. \cite{hopf}, using energy function, global stability of the fixed points is analyzed. 
Roughly speaking, the fixed points are globally stable if the connection parameters satisfy $w_{ij}=w_{ji}$ and 
$w_{ii}=0$. 
Note that the fixed points mean not only desired memories but also spurious memories. 

In the studies, global stability of 
fixed points has been studied in detail, however, BOAs to fixed points have not been studied sufficiently. 
It is not easy to analyze the BOA even in low dimensional nonlinear dynamical systems \cite{ott}. 
Therefore, precise/efficient  numerical analysis is necessary in the recurrent neural networks. 
However, as $N$ increases, the numerical analysis becomes impossible (the number of initial points is $2^N$, curse of dimensionality).

\subsection{Classification of binary data sets}
\label{class}
In order to realize efficient analysis of BOAs, we present a simple problem
: classification of binary data set. 
First, we consider a data set $U$ consisting of $N$-dimensional binary vectors
\[
U \equiv \{U_1, \cdots, U_M \},  \ 
U_i \equiv (U_{i1}, \cdots, U_{iN}), \
U_{ij} \in \{-1, +1\} 
\]
where $M$ is the number of data.  
If $M \ll 2^N$ then we can escape form the curse of dimensionality. 
A concrete example of $U$ is introduced in \ref{ab}. 

Second, we select $m$ representative vectors form the data set
\[
R_d \equiv \{ r_1, \cdots, r_m \} \subset U, \ 
r_l = (r_{l1}, \cdots, r_{lN}), \ 
r_{lj} \in \{-1, +1\}
\]
The binary vectors $r_j$ correspond to desired memories. 

Third, applying a suitable method, we try to set parameters to make BFPs without periodic orbits (oscillation). 
For example, applying the correlation based learning,  the parameters are set as the following
\begin{equation}
w_{ij} = \sum_{l=1}^m r_{li} r_{lj}, \ 
w_{ii}=0, Th=\mbox{control parameter}
\label{hebb}
\end{equation}
We consider effects of $Th$ for performance of HYN. 
If representative vectors are selected suitably, they may be stored as BFPs. 
Although guaranteed storage of all desired memories is not easy, the learning of Equation \eqref{hebb} can make several fixed points. 
We assume that the HYN has $m$ fixed points
$\bm{z}_{f1}, \cdots, \bm{z}_{fm}$. 

Forth, we apply an element $U_i \in U$ as initial point. 
If the sequence from $U_i$ falls into the $k$-th fixed point $\bm{z}_{fk}$ then $U_i$ is declared as an EFP of the $\bm{z}_k$ and classified into the $k$-th class.  
Repeating the classification for all the elements in $U$, we obtain the distribution of BOA sizes $K$ in Equation \eqref{dist}. 
Calculating normalized entropy $H$, the classification is evaluated. 

\begin{figure}[t!]
\centering
\includegraphics[width=1\columnwidth]{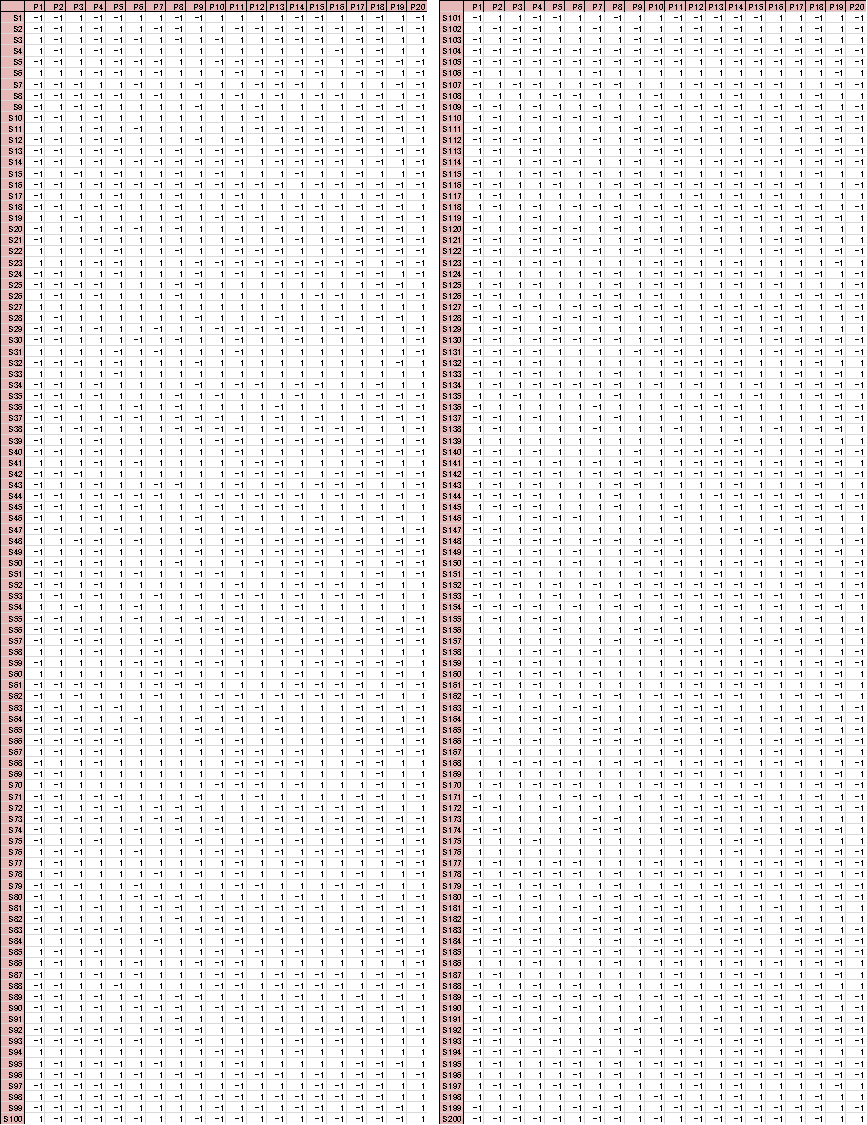}
\caption{Binary data set. Response of $N=200$ students to $M=20$ problems. }
\label{data}
\end{figure}
\section{Numerical Analysis for Basin of attraction}
Fig. \ref{data} shows binary data sets for numerical analysis of BOAs to BFPS. 
Using Gaussian random number $N(\mu, \sigma^2)$, $\mu=10, \sigma=4$, the data set is generated
\begin{equation}
U = \{U_1, \cdots, U_{200} \}, \
U_i =(U_{i1}, \cdots, U_{i20}), \ 
(N=20, M=200)
\label{rdata}
\end{equation}
\subsection{Typical example}
\label{example}
We select the following $m=4$ representative vectors from $U$: 
{\small
\[
\begin{aligned}
r_1 &= (-1,-1,-1,-1,-1,-1,-1,-1,-1,-1,-1,\phantom{-}1,\phantom{-}1,-1,-1,-1,-1,-1,-1,-1) \\
r_2 &= (-1,\phantom{-}1,-1,-1,\phantom{-}1,-1,-1,-1,-1,-1,-1,\phantom{-}1,-1,-1,-1,-1,-1,\phantom{-}1,-1,-1) \\
r_3 &= (\phantom{-}1,\phantom{-}1,\phantom{-}1,-1,\phantom{-}1,\phantom{-}1,-1,\phantom{-}1,\phantom{-}1,\phantom{-}1,-1,\phantom{-}1,\phantom{-}1,-1,\phantom{-}1,\phantom{-}1,\phantom{-}1,\phantom{-}1,\phantom{-}1,\phantom{-}1) \\
r_4 &= (\phantom{-}1,\phantom{-}1,\phantom{-}1,\phantom{-}1,\phantom{-}1,\phantom{-}1,-1,-1,\phantom{-}1,\phantom{-}1,-1,-1,-1,\phantom{-}1,\phantom{-}1,\phantom{-}1,\phantom{-}1,\phantom{-}1,\phantom{-}1,\phantom{-}1)
\end{aligned}
\]
}
Applying the correlation based learning \eqref{hebb}, 
we obtain the connection matrix 
{\scriptsize
\begin{equation}
\bm{W} =
\left(
\begin{array}{rrrrrrrrrrrrrrrrrrrr}
 0 &  2 &  4 &  2 &  2 &  4 &  0 & \phantom{-}2 &  4 &  4 &  0 & -2 &  0 &  2 &  4 &  4 &  4 &  2 &  4 &  4 \\
 2 &  0 &  2 &  0 &  4 &  2 & -2 &  0 &  2 &  2 & -2 &  0 & -2 &  0 &  2 &  2 &  2 &  4 &  2 &  2 \\
 4 &  2 &  0 &  2 &  2 &  4 &  0 &  2 &  4 &  4 &  0 & -2 &  0 &  2 &  4 &  4 &  4 &  2 &  4 &  4 \\
 2 &  0 &  2 &  0 &  0 &  2 &  2 &  0 &  2 &  2 &  2 & -4 & -2 &  4 &  2 &  2 &  2 &  0 &  2 &  2 \\
 2 &  4 &  2 &  0 &  0 &  2 & -2 &  0 &  2 &  2 & -2 &  0 & -2 &  0 &  2 &  2 &  2 &  4 &  2 &  2 \\
 4 &  2 &  4 &  2 &  2 &  0 &  0 &  2 &  4 &  4 &  0 & -2 &  0 &  2 &  4 &  4 &  4 &  2 &  4 &  4 \\
 0 & -2 &  0 &  2 & -2 &  0 &  0 &  2 &  0 &  0 &  4 & -2 &  0 &  2 &  0 &  0 &  0 & -2 &  0 &  0 \\
 2 &  0 &  2 &  0 &  0 &  2 &  2 &  0 &  2 &  2 &  2 &  0 &  2 &  0 &  2 &  2 &  2 &  0 &  2 &  2 \\
 4 &  2 &  4 &  2 &  2 &  4 &  0 &  2 &  0 &  4 &  0 & -2 &  0 &  2 &  4 &  4 &  4 &  2 &  4 &  4 \\
 4 &  2 &  4 &  2 &  2 &  4 &  0 &  2 &  4 &  0 &  0 & -2 &  0 &  2 &  4 &  4 &  4 &  2 &  4 &  4 \\
 0 & -2 &  0 &  2 & -2 &  0 &  4 &  2 &  0 &  0 &  0 & -2 &  0 &  2 &  0 &  0 &  0 & -2 &  0 &  0 \\
-2 &  0 & -2 & -4 &  0 & -2 & -2 &  0 & -2 & -2 & -2 &  0 &  2 & -4 & -2 & -2 & -2 &  0 & -2 & -2 \\
 0 & -2 &  0 & -2 & -2 &  0 &  0 &  2 &  0 &  0 &  0 &  2 &  0 & -2 &  0 &  0 &  0 & -2 &  0 &  0 \\
 2 &  0 &  2 &  4 &  0 &  2 &  2 &  0 &  2 &  2 &  2 & -4 & -2 &  0 &  2 &  2 &  2 &  0 &  2 &  2 \\
 4 &  2 &  4 &  2 &  2 &  4 &  0 &  2 &  4 &  4 &  0 & -2 &  0 &  2 &  0 &  4 &  4 &  2 &  4 &  4 \\
 4 &  2 &  4 &  2 &  2 &  4 &  0 &  2 &  4 &  4 &  0 & -2 &  0 &  2 &  4 &  0 &  4 &  2 &  4 &  4 \\
 4 &  2 &  4 &  2 &  2 &  4 &  0 &  2 &  4 &  4 &  0 & -2 &  0 &  2 &  4 &  4 &  0 &  2 &  4 &  4 \\
 2 &  4 &  2 &  0 &  4 &  2 & -2 &  0 &  2 &  2 & -2 &  0 & -2 &  0 &  2 &  2 &  2 &  0 &  2 &  2 \\
 4 &  2 &  4 &  2 &  2 &  4 &  0 &  2 &  4 &  4 &  0 & -2 &  0 &  2 &  4 &  4 &  4 &  2 &  0 &  4 \\
 4 &  2 &  4 &  2 &  2 &  4 &  0 &  2 &  4 &  4 &  0 & -2 &  0 &  2 &  4 &  4 &  4 &  2 &  4 &  0
\end{array}
\right). 
\end{equation}
}
Adjusting the threshold parameter to $Th=2.5$, the HYN has 4 fixed points:
{\small
\[
\begin{aligned}
\bm{z}_{f1} &= (-1,-1,-1,-1,-1,-1,\phantom{-}1,-1,-1,-1,\phantom{-}1,\phantom{-}1,\phantom{-}1,-1,-1,-1,-1,-1,-1,-1)\\
\bm{z}_{f2} &= (-1,-1,-1,-1,-1,-1,-1,-1,-1,-1,-1,\phantom{-}1,\phantom{-}1,-1,-1,-1,-1,-1,-1,-1)\\
\bm{z}_{f3} &= (\phantom{-}1,\phantom{-}1,\phantom{-}1,\phantom{-}1,\phantom{-}1,\phantom{-}1,\phantom{-}1,\phantom{-}1,\phantom{-}1,\phantom{-}1,\phantom{-}1,-1,-1,\phantom{-}1,\phantom{-}1,\phantom{-}1,\phantom{-}1,\phantom{-}1,\phantom{-}1,\phantom{-}1)\\
\bm{z}_{f4} &= (\phantom{-}1,\phantom{-}1,\phantom{-}1,\phantom{-}1,\phantom{-}1,\phantom{-}1,-1,\phantom{-}1,\phantom{-}1,\phantom{-}1,-1,-1,-1,\phantom{-}1,\phantom{-}1,\phantom{-}1,\phantom{-}1,\phantom{-}1,\phantom{-}1,\phantom{-}1)
\end{aligned}
\]
}
where 2 of the 4 representative vectors ($r_1$ and $r_4$) relates to the fixed points: 
\[
z_{f1} \simeq -r_4, \ 
z_{f2} = r_1, \
z_{f3} = -r_1,  \ 
z_{f4} \simeq r_4, \ 
\mbox{HD}(z_{f4}, r_4)=1. 
\]
Applying the classification method in \ref{class}, the binary data set $U$ of $M=200$ elements is classified into 4 classes with large entropy: 
%
\begin{equation}
\begin{array}{l}
K=\{k_1+1, k_2+1, k_3+1, k_4+1\} = \{51, 54, 48, 47\}\\
H=0.998 \ (p_1=\frac{51}{200}, p_2=\frac{54}{200}, p_3=\frac{48}{200}, p_4=\frac{47}{200})
\end{array}
\label{highent}
\end{equation}
%

\subsection{Classification performance}
In the numerical experiments, $10^5$ kinds of 4 representation vectors are selected from  the data set $U$. 
Applying the correlation based learning of Equation \eqref{hebb}, we obtain the connection matrix $\bm{W}$. 
For each $\bm{W}$, adjusting the threshold parameter $Th$, we have extracted the case of globally stable binary fixed points (GBFPs): all initial points fall into fixed points, no periodic orbits. 
In the $10^5$ trials, the number of such cases (GBFPs) is as the following
\[
\begin{array}{ll}
Th=0: \mbox{\#GBFPs}=7530 (7.53\%) & 
Th=0.5: \mbox{\#GBFPs}=18592 (18.6\%)\\
Th=2.5: \mbox{\#GBFPs}=67236 (67.2\%) \  &
Th=4.5: \mbox{\#GBFPs}=100000 (100\%)
\end{array}
\]
\begin{figure}[t!]
\centering
\includegraphics[width=1\columnwidth]{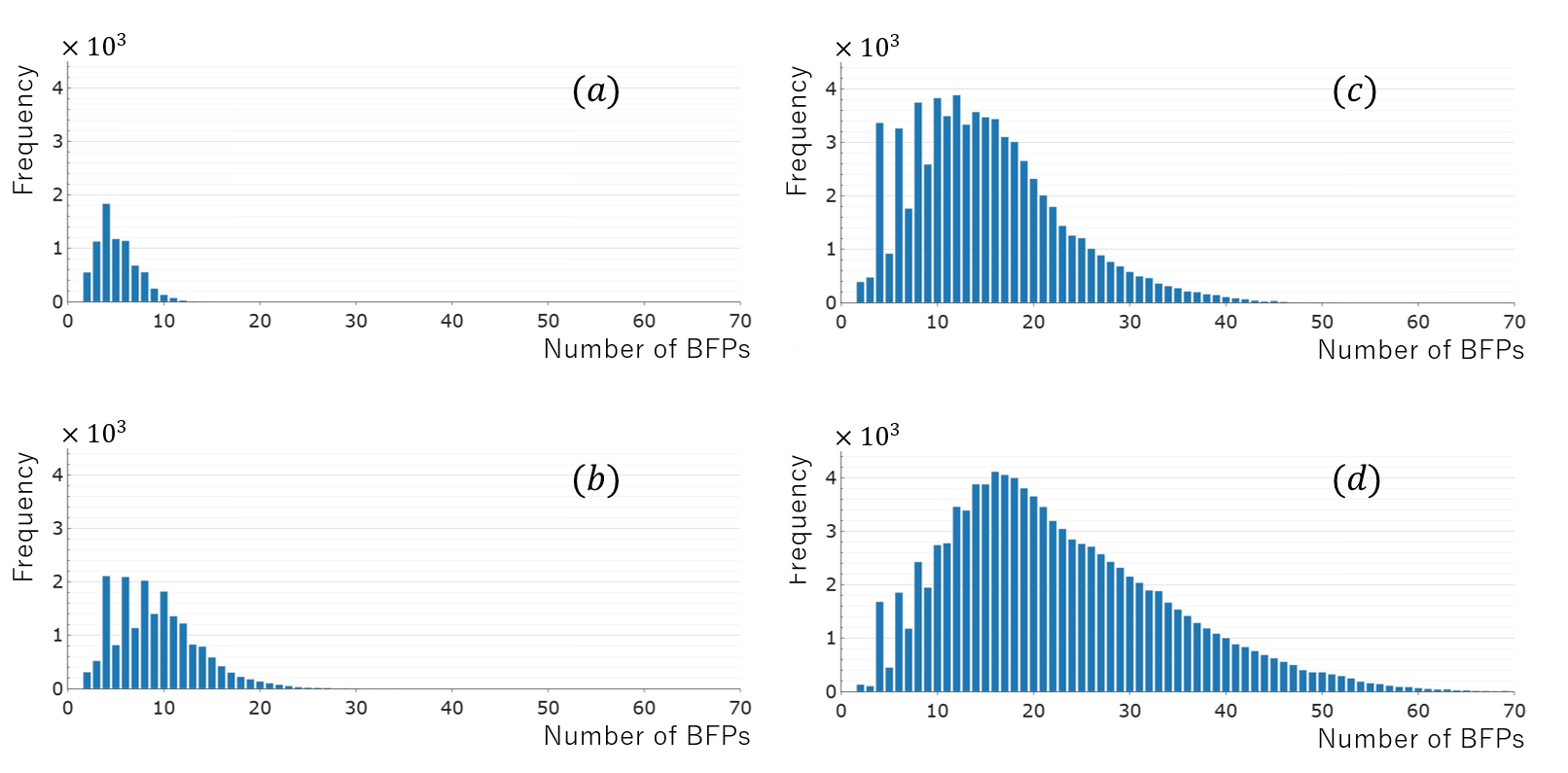}\caption{Distribution of \#BFPs in $10^5$ trials. (a) $Th=0$ (sgn). (b) $Th=0.5$ or 1.5. (c) $Th=2.5$ or 3.5. (d) $Th=4.5$.}
\label{BFP}
\centering
\includegraphics[width=1\columnwidth]{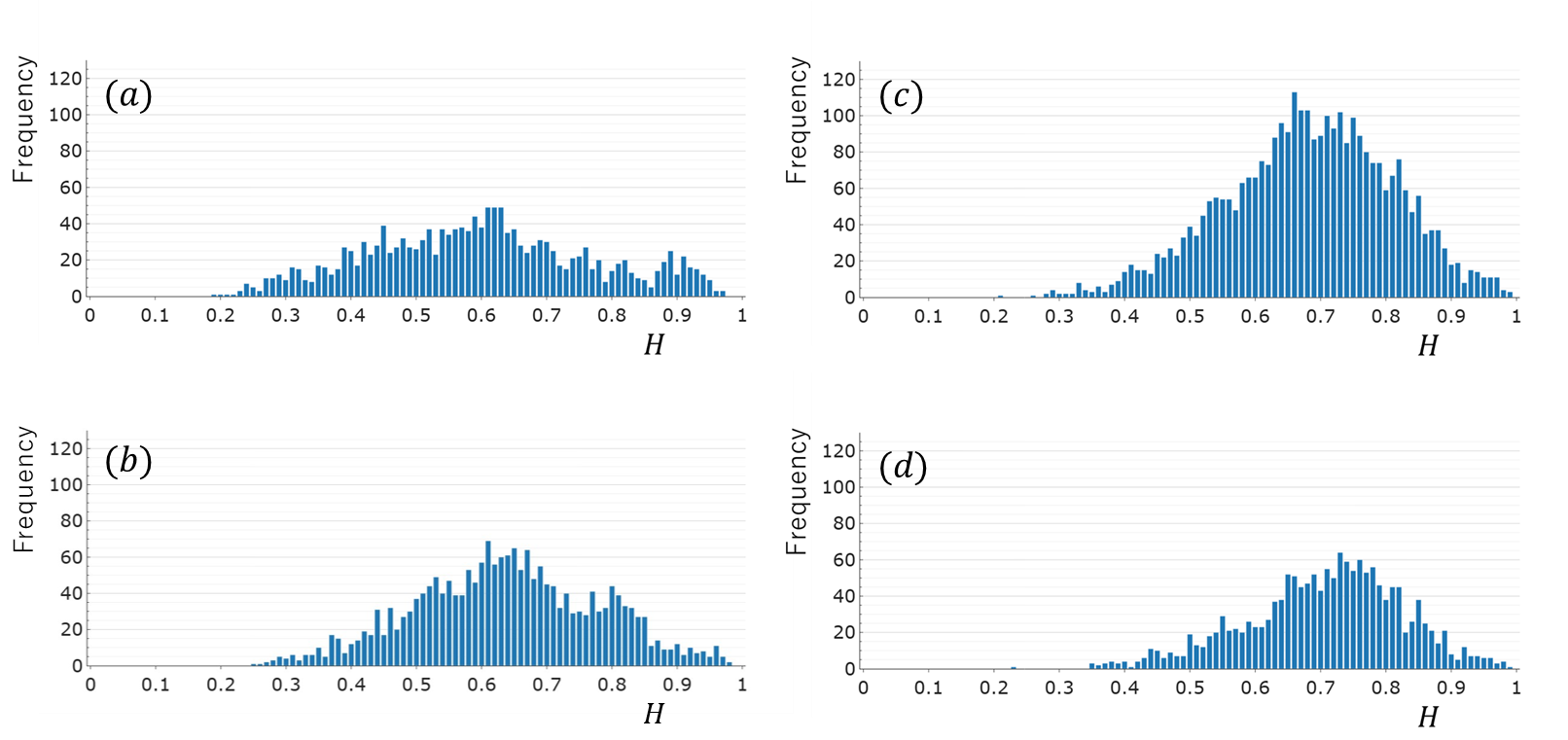}
\caption{Distribution of normalized entropy $H$ in the case of 4 BFPs. (a) $Th=0$ (sgn). (b) $Th=0.5$. (c) $Th=2.5$. (d) $Th=4.5$.}
\label{entnorm}
\end{figure}
\noindent
Since $w_{ij}$ is even, $Th \in \{0.5, 1.5 \}$ and $Th \in \{2.5, 3.5 \}$ give the same result. 
Fig. \ref{BFP} shows distribution of the number of fixed points for $Th$. 
We can see that, as $Th$ increases, distribution becomes broader. 
For simplicity, we focus on the case of 4 BFPs and have calculated EFPs: the set of BOAs of Equation \eqref{dist}.  
For the distribution of the BOA sizes,  we have calculated the normalized entropy $H$.
Fig. \ref{entnorm} shows distribution of $H$ for $Th$. 
As $Th$ increases, the entropy tend to be larger. 
In the distribution, we count the number of high entropy as summarized  in  Table \ref{tb1}. 
We can see the following:

\begin{itemize}
\item If uniform distribution of BOA sizes is required, $Th=2.5$ is suitable. 
There exists some suitable value $Th_o$ that can maximize entropy: uniform distribution of BOA sizes. 
\item If $Th$ is smaller than $Th_o$, periodic orbits are generated and  stability of BFPs becomes weak. 
The case GBFPs is hard to obtain. 
\item If $Th$ is larger than $Th_o$, the HYN has many BFPs including many undesired fixed points corresponding to spurious memories. 
\end{itemize}
We have confirmed similar results for the other data sets from 
Gaussian random number of $\mu = 10$, $\sigma = 4 \pm 0.2$.  
\begin{table}[t!]
\renewcommand{\arraystretch}{1.2}
\centering
\caption{The number of large entropy $H$ for $Th$ in the case were the HYN has 4 BFPs without periodic orbits. \#4BFPs: the number of the cases in $10^5$ trials.} 
\begin{tabular}{|c|c|c|c|c|} \hline
\ \ $Th$ \ \   & \ \  $ H>0.95 $  \ \ &  \ \ $ H>0.90 $ \ \  &  \ \ $ H>0.85 $ \ \  & \ \  \#4BFPs  \ \  \\  \hline
0   &   15   &     92   &    164 &    1620  \\ \hline
0.5  &    23    &   66    &     136   &     1994 \\ \hline
2.5   &   40    &    114   &  306   &  3163   \\ \hline
4.5  &    20    &   59    &   178    &  1568   \\ \hline
    \end{tabular}
\label{tb1}
\end{table}
\subsection{Application to binary data sets in education}
\label{ab}
The BOAs are applicable to a classification problem in education. 
In the application, the binary data set $U$ in Equation \eqref{rdata} corresponds to item response set of $M$ students.   
The $i$-th binary vector $U_i =(U_{i1}, \cdots, U_{iN})\in U$ is a response vector of the $i$-th student for $N$ problems (items): 
$U_{ij} = +1$ (respectively $-1$) means correct (respectively, incorrect) answer for the $j$-th problem. 
In the application, we try to classify the large data set $U$ (school) into subset $S_1$ to $S_m$ (classes). 
If the number of students ($M$) is large, the data set consists of a large variety of students and instruction may not be easy. 
If a suitable size of subsets is obtained, the instruction may become effective. 
In the classification, it is necessary to extract characteristics of each student. 
We try to extract feature quantities based on the 2-parameter logistic model in item response theory \cite{irt}:
\begin{equation}
P_i(\theta)=\frac{1}{1+\exp[-a_i(b_i-\theta)]}
\label{2PL}
\end{equation}
where 
$\theta$ denotes the difficulty of a problem and 
$P_i(\theta)$ denotes probability of correct answer of the $i$-th student for $\theta$. 
The parameter $a_i$ denotes discriminating power of the $i$-th student and 
$b_i$ denotes ability of the $i$-th student. 
$a_i$ and $b_i$ are used to characterize the $i$-th student. 

\begin{algorithm}
    \caption{Extraction of feature quantities $a_i$ and $b_i$}
    \label{alg1}
    \begin{algorithmic}
        \STATE $U \in \{0,1\}^{M \times N}$  (element $U_{ij} \leftarrow (U_{ij} + 1 )/2$) \hfill //Binary data set
        \STATE Discretize $\theta$ into $K$ equally spaced latent points $\theta_1,\ldots,\theta_K$ on $[-4,4]$ 
        \FOR{$k = 1$ to $K$}
            \STATE   $g(\theta_k)\leftarrow\dfrac{\phi(\theta_k)}{\sum_{k'=1}^{K}\phi(\theta_{k'})}$
            \hfill //Set prior distribution $g(\theta_k)$
            \ENDFOR
        \FOR{$i = 1$ to $M$} 
            \STATE $a_i \leftarrow 1$, \; $b_i \leftarrow 0$ 
            \hfill //Initial value
            \ENDFOR 
            \REPEAT 
                \STATE 
                \textbf{E-step} 
                \FOR{$k = 1$ to $K$} 
                    \FOR{$i = 1$ to $M$} 
                        \STATE 
                        $ P_i(\theta_k) \leftarrow \dfrac{1} { 1+\exp[-a_i(b_i-\theta_k)] } $ 
                        \hfill //Logistic model
                    \ENDFOR 
                    \FOR{$j = 1$ to $N$} 
                        \STATE 
                        $ L_{jk} \leftarrow \displaystyle \prod_{i=1}^{M} P_i(\theta_k)^{U_{ij}} (1-P_i(\theta_k))^{1-U_{ij}} $ 
                        \hfill //Likelihood of $\theta_k$
                    \ENDFOR 
                \ENDFOR 
                \STATE $u_j=(U_{1j},U_{2j},\ldots,U_{Mj})^{\mathrm T}$
                \hfill //Response vector of item $j$
                \FOR{$j = 1$ to $N$} 
                    \FOR{$k = 1$ to $K$} 
                        \STATE $ p(\theta_k|u_j) \leftarrow L_{jk} g(\theta_k) $ 
                    \ENDFOR 
                    \STATE Normalize: $ p(\theta_k|u_j) \leftarrow \dfrac{ p(\theta_k|u_j) }{ \sum_{k'=1}^{K} p(\theta_{k'}|u_j) } $
                    \hfill //Posterior probability
                \ENDFOR 
                \STATE \textbf{M-step} 
                \FOR{$i = 1$ to $M$} 
                    \STATE Update $(a_i,b_i)$ by maximizing 
                    \STATE \[ \sum_{j=1}^{N} \sum_{k=1}^{K} p(\theta_k|u_j) \left[ U_{ij}\log P_i(\theta_k) + (1-U_{ij}) \log(1-P_i(\theta_k)) \right] \] 
                    \STATE Optimize $(a_i,b_i)$with bounds\qquad
                    $a_i \in [0.01,3],\; b_i \in [-4,4]$
                    \hfill // Estimate $a_i, b_i$
                \ENDFOR 
            \UNTIL{convergence} 
            
    \end{algorithmic}
\end{algorithm}
\clearpage
\begin{figure}[t!]
\centering
\includegraphics[width=0.5\columnwidth]{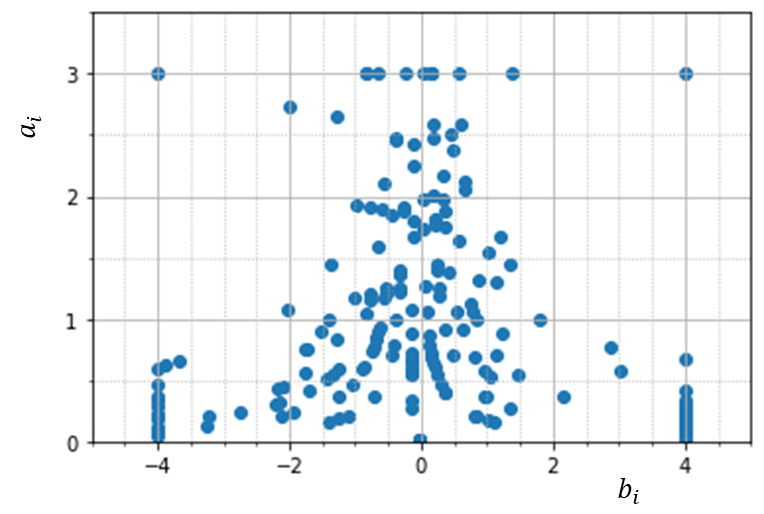}
\caption{Feature quantities for all students ($M=200$). 
$a_i$: discriminating power.  
$b_i$: ability. $i \in \{1, \cdots, 200 \}$
}
\label{all}

\vspace*{3mm}

\centering
\includegraphics[width=0.95\columnwidth]{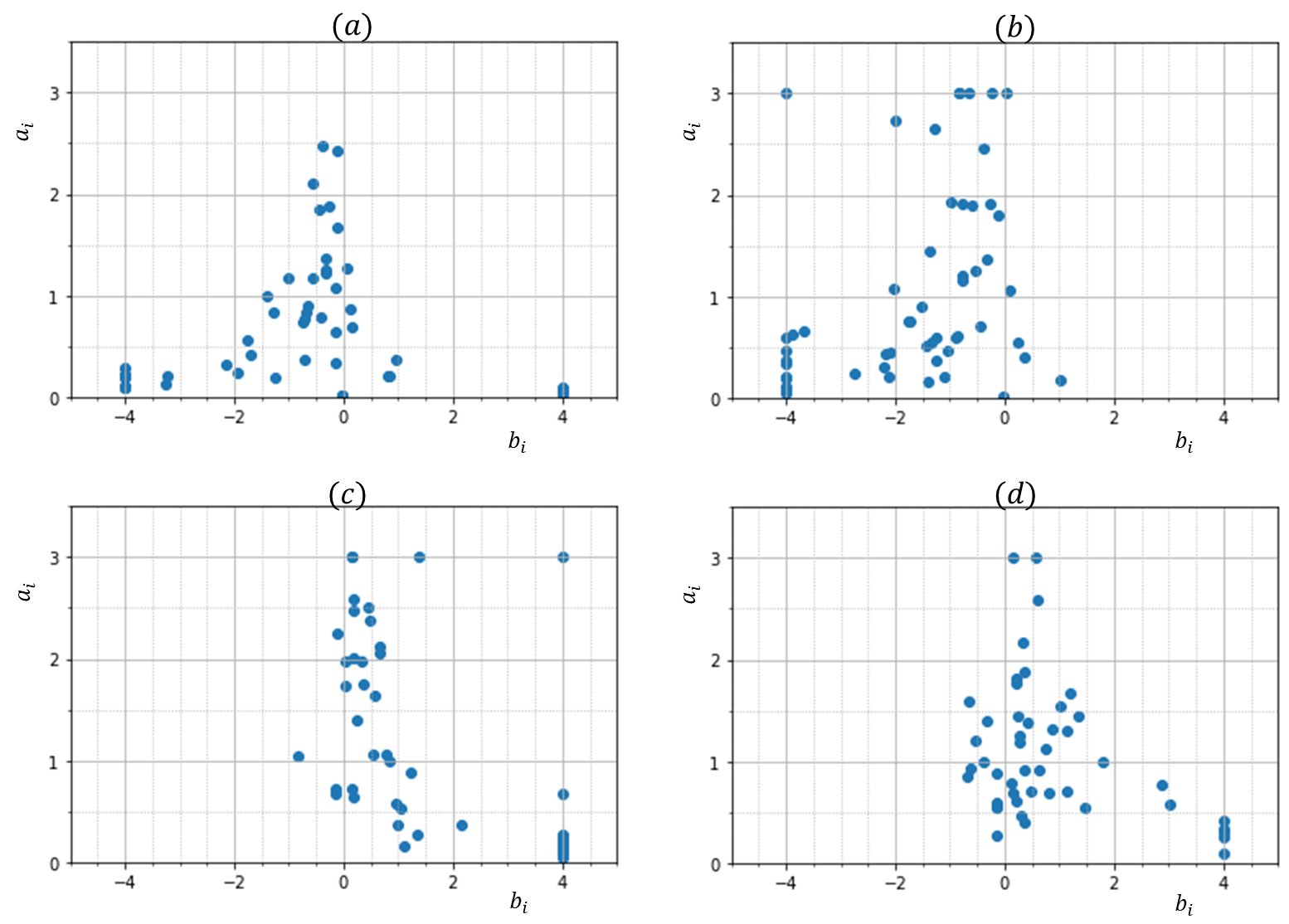}
\caption{Feature quantities for 4 classes. (a) Class 1 of 51 students: low $b_i$, low $a_i$. (b) 54 students: low $b_i$, wide $a_i$.(c) 48 students: medium $b_i$, wide $a_i$. (d) 47 students: high $b_i$, low $a_i$. }
\label{4cls}
\end{figure}
%
In order to estimate the feature quantities $a_i$ and $b_i$, we use the standard maximum likelihood routine. 
Algorithm \ref{alg1} shows the quasi code where $U_{ij}$ is the response of the $i$-th student to the $j$-th problem (item) and is summarized into the data set $U$. 
Fig. \ref{all} shows distribution of the feature quantities for $M=200$ students in the binary data set in Fig. \ref{data}. 
The quantities are distributed in wide area, 
corresponding to a variety of students. 
Applying the classification method in subsection \ref{class}, 
we obtain 4 classes with large entropy as shown in subsection \ref{example}. 
Fig. \ref{4cls} shows distributions of the 4 classes. 
Each class seems to have inherent property. 
Depending on features of the item response data set, the BOA based method can realize effective classification of students. 

\section{Conclusions}
BOAs to multiple fixed points has been studied in the HYN. 
In order to escape from the curse of dimensionality, we present a classification problem of binary data sets. 
In order to evaluate the distribution of BOA sizes, we use the normalized entropy. 
Applying a correlation based learning, we obtain connection parameters that make fixed points in the HYN. 
Performing precise numerical analysis, we have clarified that the threshold parameter can maximize the entropy: the distribution of basins approaches to be uniform. 
A fundamental step to application to item response data sets is also considered.   

Future problems are many, including the following. 
Detailed analysis of BOAs in various learning methods, 
theoretical analysis of the distribution of BOA sizes, and 
application to classification of various data sets.  

\section*{Acknowledgment}
The authors wish to thank Prof. Kenya Jin'no at Tokyo city university for his valuable advices on HYN and  classification.

\end{document}